\documentclass[11pt]{article}
\usepackage{coling2020}
\usepackage{times}
\usepackage{url}
\usepackage{latexsym}

\usepackage{multirow}
\usepackage{booktabs}

\usepackage{todonotes}
\newcommand{\SideNote}[2]{} 
\renewcommand{\SideNote}[2]{\todo[color=#1,size=\small]{#2}} %

\usepackage{float}
\newfloat{floatquote}{tbp}{loq}
\floatname{floatquote}{Quote}
\usepackage{caption}
\usepackage{cleveref}
\crefname{floatquote}{quote}{quotes}
\Crefname{floatquote}{Quote}{Quotes}
\usepackage{enumitem}
\usepackage{subcaption}
\usepackage{linguex}
\newcommand*\rot{\rotatebox{60}}

\usepackage{hhline}
\usepackage{tabularx}

\title{Querent Intent in Multi-Sentence Questions}

\author{Laurie Burchell\thanks{~Alphabetical order, equal contribution}, Jie Chi\footnotemark[1], Tom Hosking\footnotemark[1], Nina Markl\footnotemark[1], Bonnie Webber \\ 
  Institute for Language, Cognition and Computation \\ University of Edinburgh \\
  {\tt \{laurie.burchell,jie.chi,tom.hosking,nina.markl\}@ed.ac.uk} \\
  }

\date{}

\begin{document}
\maketitle

\begin{abstract}

Multi-sentence questions (MSQs) are sequences of questions connected by relations which, unlike sequences of standalone questions, need to be answered as a unit. Following Rhetorical Structure Theory (RST), we recognise that different ``question discourse relations'' between the subparts of MSQs reflect different speaker intents, and consequently elicit different answering strategies. Correctly identifying these relations is therefore a crucial step in automatically answering MSQs. We identify five different types of MSQs in English, and define five novel relations to describe them. We extract over 162,000 MSQs from Stack Exchange to enable future research. Finally, we implement a high-precision baseline classifier based on surface features.

\end{abstract}

%
%
\blfootnote{
    %
    %
    %
    %
    \hspace{-0.65cm}  
    This work is licensed under a Creative Commons 
    Attribution 4.0 International Licence.
    Licence details:
    \url{http://creativecommons.org/licenses/by/4.0/}.
    %
    %
}

\section{Introduction}
\label{sec:intro}



A multi-sentence question (MSQ) is a dialogue turn that contains more than one question (cf. Ex. \ref{ex1}). We refer to the speaker of such a turn as a \textit{querent} (i.e., one who seeks).

\ex. \a.[Querent: ] How can I transport my cats if I am moving a long distance? (Q1) 
    \b.[]      For example, flying them from NYC to London? (Q2) \label{ex1}

A standard question answering system might consider these questions separately:  

\ex. \label{answer1} \a.[A1: ] You can take them in the car with you. 
    \b.[A2: ] British Airways fly from NYC to London.

However, this naïve approach does not result in a good answer, since the querent intends that an answer take both questions into account: in \ref{ex1}, Q2 clarifies that taking pets by car is not a relevant option. The querent is likely looking for an answer like \ref{answer2}: 

\ex.\label{answer2} \a.[A: ] British Airways will let you fly pets from NYC to London.

Whilst question answering (QA) has received significant research attention in recent years \cite{TriviaQA,VisualQA}, there is little research to date on answering MSQs, despite their prevalence in English. Furthermore, existing QA datasets are not appropriate for the study of MSQs as they tend to be sequences of standalone questions constructed in relation to a text by crowdworkers (e.g. SQuAD \cite{rajpurkar2016squad}). We are not aware of any work that has attempted to improve QA performance on MSQs, despite the potential for obvious errors as in the example above.

Our contribution towards the broader research goal of automatically answering MSQs is as follows:
\vspace{-0.5em}
\begin{itemize}
    \setlength\itemsep{-0.2em}
    \item We create a new dataset of 162,745 English two-question MSQs from Stack Exchange.
    \item We define five types of MSQ according to how they are intended to be answered, inferring intent from relations between them. 
    \item We design a baseline classifier based on surface features. 
\end{itemize}

\section{Prior work}
\label{sec:prior}

Prior work on QA has focused on either single questions contained within dialogue \cite{choi2018quac,reddy-etal-2019-coqa,sharc,clark2018think}, or questions composed of two or more sentences crowd-sourced by community QA (cQA) services \cite{issuesincqa,cqamsq}. Our definition of MSQs is similar to the latter, but it should be noted that sentences in existing cQA datasets can be declarative or standalone, while in our case they must be a sequence of questions that jointly imply some user intent. Popular tasks on cQA have only considered the semantics of individual questions and answers, while we are more focused on interactions between questions.

\newcite{huang-etal-2008-question} and \newcite{anstypeinf} classify questions to improve QA performance, but their work is limited to standalone questions. \newcite{ciurca2019} was the first to identify MSQs as a distinct phenomenon, and curated a small dataset consisting of 300 MSQs extracted from Yahoo Answers. However, this dataset is too small to enable significant progress on automatic classification of MSQ intent.



\section{Large-scale MSQ dataset}
\label{sec:data}

Stack Exchange is a network of question-answering sites, where each site covers a particular topic. Questions on Stack Exchange are formatted to have a short title and then a longer body describing the question, meaning that it is far more likely to contain MSQs than other question answering sites, which tend to focus attention on the title with only a short amount of description after the title. There is a voting system which allows us to proxy well-formedness, since badly-formed questions are likely to be rated poorly. It covers a variety of topics, meaning that we can obtain questions from a variety of domains.

To obtain the data, we used the Stack Exchange Data Explorer\footnote{https://data.stackexchange.com/}, an open source tool for running arbitrary queries against public data from the Stack Exchange network. We chose 93 sites within the network, and queried each site for entries with at least two question marks in the body of the question. We removed any questions with \TeX{} and mark-up tags, then replaced any text matching a RegEx pattern for a website with `\texttt{[website]}'. From this cleaned text, we extracted pairs of MSQs by splitting the cleaned body of the question into sentences, then finding two adjacent sentences ending in `\texttt{?}'. We removed questions under 5 or over 300 characters in length. Finally, we removed any question identified as non-English using \texttt{langid.py} \cite{lui-baldwin-2012-langid}. Many of the questions labelled as `non-English' were in fact badly formed English, making language identification a useful pre-processing step.

After cleaning and processing, we extracted 162,745 questions from 93 topics\footnote{Our dataset is available at \url{https://github.com/laurieburchell/multi-sentence-questions}}. A full list of topics and the number of questions extracted from each is given in \Cref{app:topics}. We restrict the dataset to pairs of questions, leaving longer sequences of MSQs for future work.



\begin{table}[!ht]
    \centering
    \newcolumntype{L}{>{\raggedright\arraybackslash}X}%

    \begin{tabularx}{\textwidth}{rL}
        \multicolumn{2}{l}{\textsc{Separable}} \\
        \hhline{==}
        \textbf{Example} & \textit{What’s the recommended kitten food?} \newline \textit{How often should I feed it?} \\
        \textbf{Intent} & Two questions on the same topic (the querent’s kitten). \\
        \textbf{Strategy} & Resolve coreference and answer both questions separately. \\
        
        \multicolumn{2}{l}{} \\
        
        \multicolumn{2}{l}{\textsc{Reformulated}} \\
        \hhline{==}
        \textbf{Example} & 
 \textit{Is Himalayan pink salt okay to use in fish tanks?} \newline \textit{I read that aquarium salt is good but would pink salt work?} \\
        \textbf{Intent} & Speaker wants to paraphrase Q1 (perhaps for clarity). \\
        \textbf{Strategy} & Answer one of the two questions. \\
        
        \multicolumn{2}{l}{} \\
        
        \multicolumn{2}{l}{\textsc{Disjunctive}} \\
        \hhline{==}
        \textbf{Example} & 
 \textit{Is it normal for my puppy to eat so quickly?} \newline \textit{Or should I take him to the vet?} \\
        \textbf{Intent} & Querent offers two potential answers in the form of polar questions. \\
        \textbf{Strategy} & Select one of the answers offered (e.g. ``Yes, it is normal") or reject both (e.g. ``Neither -- try feeding it less but more often").\\
        
        \multicolumn{2}{l}{} \\
        
        \multicolumn{2}{l}{\textsc{Conditional}} \\
        \hhline{==}
        \textbf{Example} & 
 \textit{Has something changed that is making cats harder to buy?} \newline \textit{If so, what changed?} \\
        \textbf{Intent} & Q2 only matters if the answer to Q1 is ``yes''.\\
        \textbf{Strategy} & First consider what the answer to Q1 is and then answer Q2. \\
        
        \multicolumn{2}{l}{} \\
        
        \multicolumn{2}{l}{\textsc{Elaborative}} \\
        \hhline{==}
        \textbf{Example} & 
 \textit{How can I transport my cats if I am moving a long distance?} \newline \textit{For example, flying them from NYC to London?} \\
        \textbf{Intent} & Querent wants a more specific answer. \\
        \textbf{Strategy} & Combine context and answer the second question only. \\

    \end{tabularx}
    \caption{The five types of MSQ we describe, an example of each, the querent's intent, and the resulting answering strategy. \newcite{mann1988rhetorical}'s \textsc{elaboration}, \textsc{condition} and \textsc{restatement} relations correspond roughly to three of the relations we recognise.}
    \label{tab:types}
\end{table}

\section{MSQ type as a proxy for speaker intent}
\label{sec:msq type}

MSQs are distinct from sequences of standalone questions in that their subparts need to be considered as a unit (see \ref{ex1} in Section \ref{sec:intro}). This is because they form a \textbf{discourse}: a coherent sequence of utterances \cite{Hobbs1979}. In declarative sentences, the relationship between their different parts is specified by ``discourse relations'' \cite{Stede2011,Kehler2006}, which may be signalled with discourse markers (e.g. \textit{if}, \textit{because}) or discourse adverbials  (e.g. \textit{as a result}, see \newcite{rohde2015}). We propose adapting the notion of discourse relations to interrogatives.  

A particularly useful approach to discourse relations in the context of MSQs is Rhetorical Structure Theory (RST) \cite{mann1988rhetorical}, which understands them to be an expression of the speaker's communicative intent. Listeners can infer this intent under the assumptions that speakers are ``cooperative'' and keep their contributions as brief and relevant as possible \cite{Grice1975}. Transposing this theory to interrogatives, we can conceptualise the querent's communicative intent as a specific kind of answer. Reflecting this intent, the relation suggests an answering strategy. 




We introduce five types of ``question discourse relations'' with a prototypical example from our data set, highlighting the inferred intent and the proposed answering strategy in \Cref{tab:types}.

\section{Classification using contrastive features}


Since \newcite{ciurca2019} found that using conventional discourse parsers created for declaratives is not suitable for extracting discourse relations from MSQs, we design our own annotation scheme and use it to implement a baseline classifier. Following previous work on extracting discourse relations \cite{rohde2015}, we use discourse markers and discourse adverbials alongside other markers indicative of the structure of the question (listed in \cref{app:feats}) to identify explicitly signalled relations.\footnote{Since implicit discourse relations are pervasive and challenging to automatic systems \cite{Sporleder2008}, we make no attempt to extract them here.}

We construct a high-precision, low-recall set of rules to distinguish the most prototypical forms of the five types using combinations of binary contrastive features. To derive the relevant features, we consider the minimal edits to examples of MSQs required to break or change the type of discourse relation between their parts. We then define a feature mask for each MSQ type which denotes whether each feature is required, disallowed or ignored by that type. Each mask is mutually exclusive by design.

Given a pair of questions, the system enumerates the values of each feature, and compares to the definitions in \Cref{app:mask}. If a match is found, the pair is assigned the corresponding MSQ label, otherwise it is assigned \textsc{Unknown}. This process is illustrated in \Cref{app:visual}.

\begin{figure*}[!t] 
    \begin{subfigure}[t]{.5\textwidth}
    \centering
        \includegraphics[width=\linewidth]{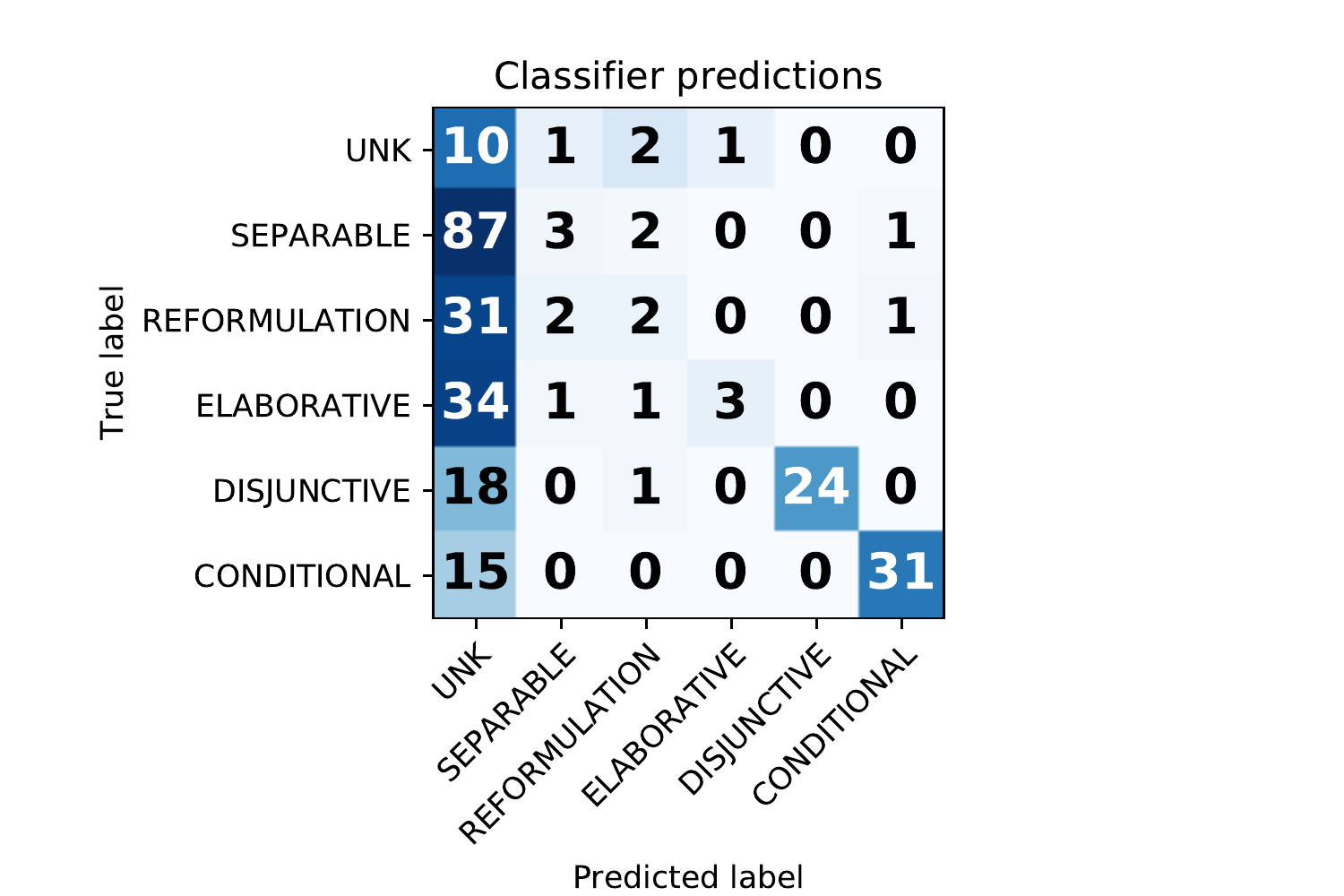}%
        
        \caption{Confusion matrix for our classifier evaluated on our hand-annotated test set. The classifier can reliably detect \textsc{Disjunctive} and \textsc{Conditional} MSQs, achieving a high overall precision score of 82.9\%.}
        \label{fig:cm_300}%
    \end{subfigure}
    ~
    \begin{subfigure}[t]{.5\textwidth}
    \centering
        \includegraphics[width=\linewidth]{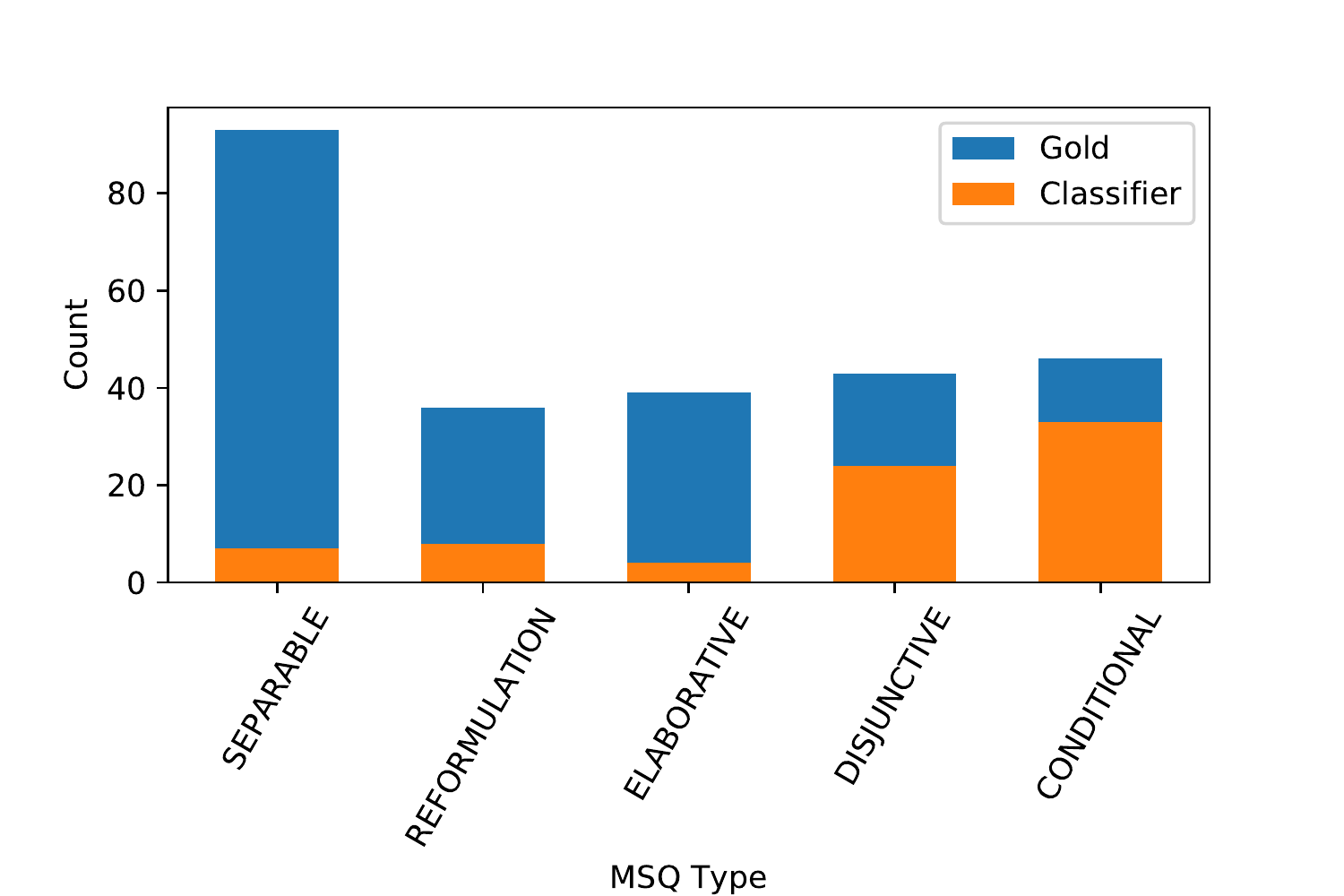}%
        
        \caption{Counts of each MSQ type in our test set, according to our annotation and our classifier. While \textsc{Separable} MSQs appear to be the most prevalent, the classifier identifies only a small fraction of them, implying that they are likely to be implicitly signalled. \textsc{Disjunctive} and \textsc{Conditional} are the most likely to be explicitly signalled.}
        \label{fig:type_counts}%
    \end{subfigure}
    \caption{}
\end{figure*}

To evaluate our classifier, 420 MSQs from our test set were annotated by two native speakers. We then evaluate the classifier on the subset of 271 samples for which both annotators agreed on the MSQ type. The resultant confusion matrix is shown in \Cref{fig:cm_300}, with the classifier achieving 82.9\% precision and 26.5\% recall.  

Overall, we find that our classifier performs well for a heuristic approach, but that real world data contains many subtleties that can break our assumptions. During the annotation process, we found many instances of single questions followed by a question which  fulfils a purely social function, such as ``\textit{Is it just me or this a problem?}'' (a \textit{phatic} question, see \newcite{Robinson1992}). MSQs can also exhibit more than one intent, presenting a challenge for both our classifier and the expert annotators (see \Cref{app:annotator}).

A limitation of our classifier is the focus on explicit MSQs, which can be identified with well-defined features. The low recall of our classifier indicates that MSQs are often implicit, missing certain markers or not completely fulfilling the distinguishing requirements. \Cref{fig:type_counts} shows that while \textsc{Disjunctive} and \textsc{Conditional} MSQs are often explicitly signalled, the other types are likely to be implicit.

\section{Conclusion}
\label{sec:conclusion}
Inspired by the role of discourse relations in MSQ answering strategies, we propose a novel definition of five different categories of MSQs based on their corresponding speaker intents. We introduce a rich and diversified multi-sentence questions dataset, which contains 162,000 MSQs extracted from Stack Exchange. This achieves our goal of providing a resource for further study of MSQs. Additionally, we implement a baseline classifier based on surface features as a prelimininary step towards successful answering strategies for MSQs.

Future work could improve on our classifier by considering implicit MSQs, with one potential approach being to transform explicit MSQs into implicit examples by removing some markers while ensuring the relation is still valid. Other areas for further work include implementing appropriate answering strategies for different types of MSQs, and investigating whether and how longer chains of MSQs differ compared to pairs of connected questions.

\section*{Acknowledgments}

This work was supported in part by the UKRI Centre for Doctoral Training in Natural Language Processing, funded by the UKRI (grant EP/S022481/1) and the University of Edinburgh. We would like to thank Bonnie Webber for her supervision, and Ivan Titov and Adam Lopez for their useful advice.

\bibliographystyle{acl}
\bibliography{coling2020}

\appendix




\onecolumn
\section{Number of questions by topic}\label{app:topics}
\begin{table}[h]
\small
    \centering
    \begin{tabular}{c|c| |c|c}
    \textbf{Topic} & \textbf{Number of questions} & \textbf{Topic} & \textbf{Number of questions} \\
    \hline
SuperUser & 17197 & Philosophy & 799 \\
ServerFault & 14780 & Hinduism & 760 \\
ElectricalEngineering & 9847 & PhysicalFitness & 760 \\
Arquade & 9704 & Homebrewing & 671 \\
Physics & 8192 & QuantFinance & 667 \\
English & 7851 & Outdoors & 599 \\
EnglishLearners & 5922 & Sports & 576 \\
SciFiFantasy & 4920 & Islam & 524 \\
InformationSecurity & 4652 & BiblicalHermeneutics & 507 \\
RPG & 4259 & Buddhism & 492 \\
DIY & 3017 & Engineering & 492 \\
Travel & 2985 & Linguistics & 484 \\
Academia & 2488 & TheorecticalCompSci & 484 \\
PersonalFinance & 2312 & Chess & 483 \\
StackOverflowMeta & 2265 & SoundDesign & 482 \\
SeasonedAdvice & 2234 & Pets & 473 \\
UX & 2219 & Economics & 472 \\
Workplace & 2189 & Parenting & 472 \\
Photography & 2111 & CognitiveSciences & 444 \\
Aviation & 1949 & Monero & 425 \\
Biology & 1832 & Health & 420 \\
Bitcoin & 1806 & ComputationalScience & 398 \\
Worldbuilding & 1801 & ReverseEngineering & 392 \\
MiYodeya & 1769 & EarthScience & 386 \\
Chemistry & 1652 & ProjectManagement & 383 \\
Music & 1589 & ArtificialIntelligence & 360 \\
ComputerScience & 1500 & Expatriates & 310 \\
Cryptography & 1487 & Robotics & 295 \\
GraphicDesign & 1450 & AmateurRadio & 285 \\
Ethereum & 1415 & Woodworking & 283 \\
BoardGames & 1402 & Literature & 234 \\
SpaceExploration & 1354 & HistoryScienceMathematics & 229 \\
Motors & 1285 & Tor & 219 \\
Bicycles & 1241 & Lego & 213 \\
Anime & 1208 & MartialArts & 202 \\
Law & 1188 & Mythology & 174 \\
NetworkEngineering & 1160 & InterpersonalSkills & 156 \\
Christianity & 1084 & Freelancing & 148 \\
Politics & 1074 & Poker & 145 \\
History & 1066 & Genealogy & 137 \\
Gardening & 1007 & MusicFans & 129 \\
DataScience & 955 & ArtsAndCrafts & 118 \\
SignalProcessing & 912 & Movies & 118 \\
Puzzling & 888 & SustainableLiving & 109 \\
Astronomy & 816 & OpenData & 105 \\
Skeptics & 807 & LifeHacks & 83 \\
Writers & 805 &  & \\
    
    \end{tabular}
    \caption{Number of questions extracted by site topic in StackOverflow dataset, sorted by descending number of questions}
    \label{tab:topics}
\end{table}

\clearpage

\section{Contrastive features}
\label{app:mask}

Note that these requirements do not apply in all cases: if all conditions are met then we assert that a pair of questions \textit{must} be of that type, but the absence of a feature does not \textit{forbid} that relation from being present. A list of the lexical markers used to define features is given in \Cref{app:feats}. Like discourse relations in declaratives, question discourse relations may be \textit{implicit}, i.e. not marked with a connective or other marker but inferable to the listener. These implicit relations continue to be very challenging for automatic systems \cite{Sporleder2008} and we do not attempt to handle them. 

\begin{table*}[ht]
    \centering
    \begin{tabular}{ll|cccccc}
         \rot{\textbf{Feature}} & \rot{\textbf{Surface} } & \rot{\textbf{Separable}} & \rot{\textbf{Reformulated}} & \rot{\textbf{Disjunctive}} & \rot{\textbf{Conditional}} & \rot{\textbf{Elaborative}} & \rot{\textbf{Elab. Statement}}\\ \toprule 
        
        \multirow{2}{*}{Anaphora} & Pronoun in Q2 & + & - & & & + & \\
        & VP ellipsis in Q2 & -  & & & & & \\ \midrule
        \multirow{5}{*}{Polarity} & Polar Q1 & - & & + & + & &  \\
        & Polar Q2 & & & + & & & - \\
        & Wh Q1 & & & - & - &  & \\
        & Wh Q2 & & & - & & - & - \\ 
        & Q2 = Statement & - & - & - & & - & +\\ \midrule
        \multirow{4}{*}{Disc. Marker} &``if" marker & - & - & - & + & - & - \\ 
        & ``or" & - & - & + & - & - & - \\
        & ``elab.'' marker & - & - & - & - & + & \\ 
        & ``sep.'' marker &  & - & - & - & - & - \\ \midrule
        Semantics & Word vector &  & + &  &  &  & + \\ 
        \end{tabular}
    \caption{Definitions for each MSQ type. `+' indicates that a feature is \textit{required}, while `-' means the feature is \textit{disallowed}. Types can be ignored for some features, meaning that the features are neither disallowed nor required.}
    \label{tab:features}
    
\end{table*}

We include the case where Q2 is a statement, as in ``\textit{How can I transport my cats if I am moving a long distance? For example, to London.}". Although these forms of MSQ do not appear in our dataset due to the filtering method used, they are in general valid, and we include them for completeness.

To evaluate semantic similarity between Q1 and Q2, we calculated the cosine similarity between the mean of the words vectors, and compared to a threshold of 0.8.

\section{Lexical Markers}
\label{app:feats}
Some of the discourse markers are drawn from the Penn Discourse Tree Bank (PDTB) annotation scheme \cite{webber2019penn}. 

\subsection{Anaphora Markers} 
To identify cases of anaphora, we searched for the following pronoun strings (and \textit{it's}) in the second question:

    she,
    he,
    it,
    they,
    her,
    his,
    its,
    their,
    them,
    it's

\subsection{Verb Ellipsis Markers} 
To identify cases of verb ellipsis, we searched for the following pro-forms of full verb phrases in the second question:

    do so,
    did so,
    does so,
    do it,
    do too,
    does too,
    did too,
    did it too,
    do it too,
    does it too

\subsection{Polar Question Markers} 

do,
	does,
	did,
	didn’t,
	will,
	won’t,
	would,
	is,
	are,
	were,
	weren’t,
	wasn’t,
	can,
	can’t,
	could,
	must,
	have,
	has,
	had,
	hasn’t,
	haven’t,
	should,
	shouldn’t,
	may,
	might,
	shall,
	ought

\subsection{Wh- words} 

    who,
    what,
    where,
    when,
    why,
    how,
    which

\subsection{Conditional (\textit{``if''}) Markers} 

    if so,
    accordingly,
    then,
    as a result,
    it follows,
    subsequently,
    consequently,
    if yes,
    if not,
    if the answer is yes,
    if the answer is no
    
\subsection{Elaborative Markers} 

    for instance,
    for example,
    e.g.,
    specifically,
    particularly,
    in particular,
    more specifically,
    more precisely,
    therefore

\subsection{Separable Markers} 

also, secondly, next, related, relatedly, similarly, furthermore

\section{Classifier visualisation}
\label{app:visual}

\begin{figure*}[h!]
    \centering
    \includegraphics[width=\textwidth]{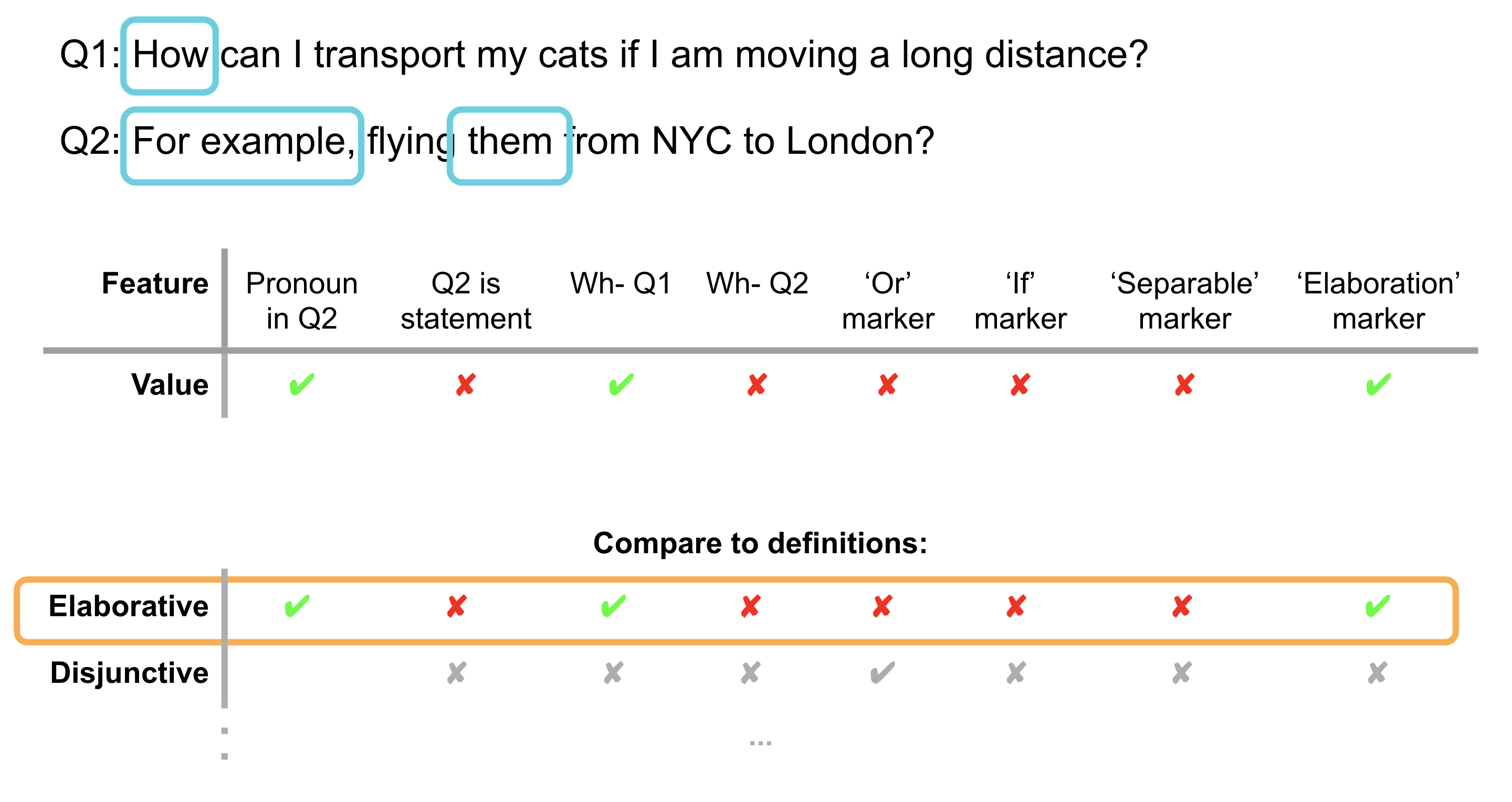}
    \caption{A visualisation of the labelling process. The system checks for the presence of each feature in the input text (circled in blue) and constructs a feature vector for the pair of questions. This feature vector is compared to the definitions, and if a match is found the questions are assigned the corresponding label. Note that not all features are shown.}
    \label{fig:worked_ex}
\end{figure*}

\clearpage

\section{Annotator agreement}
\label{app:annotator}

\begin{figure}[ht] 
    \centering
        \includegraphics[width=0.6\linewidth]{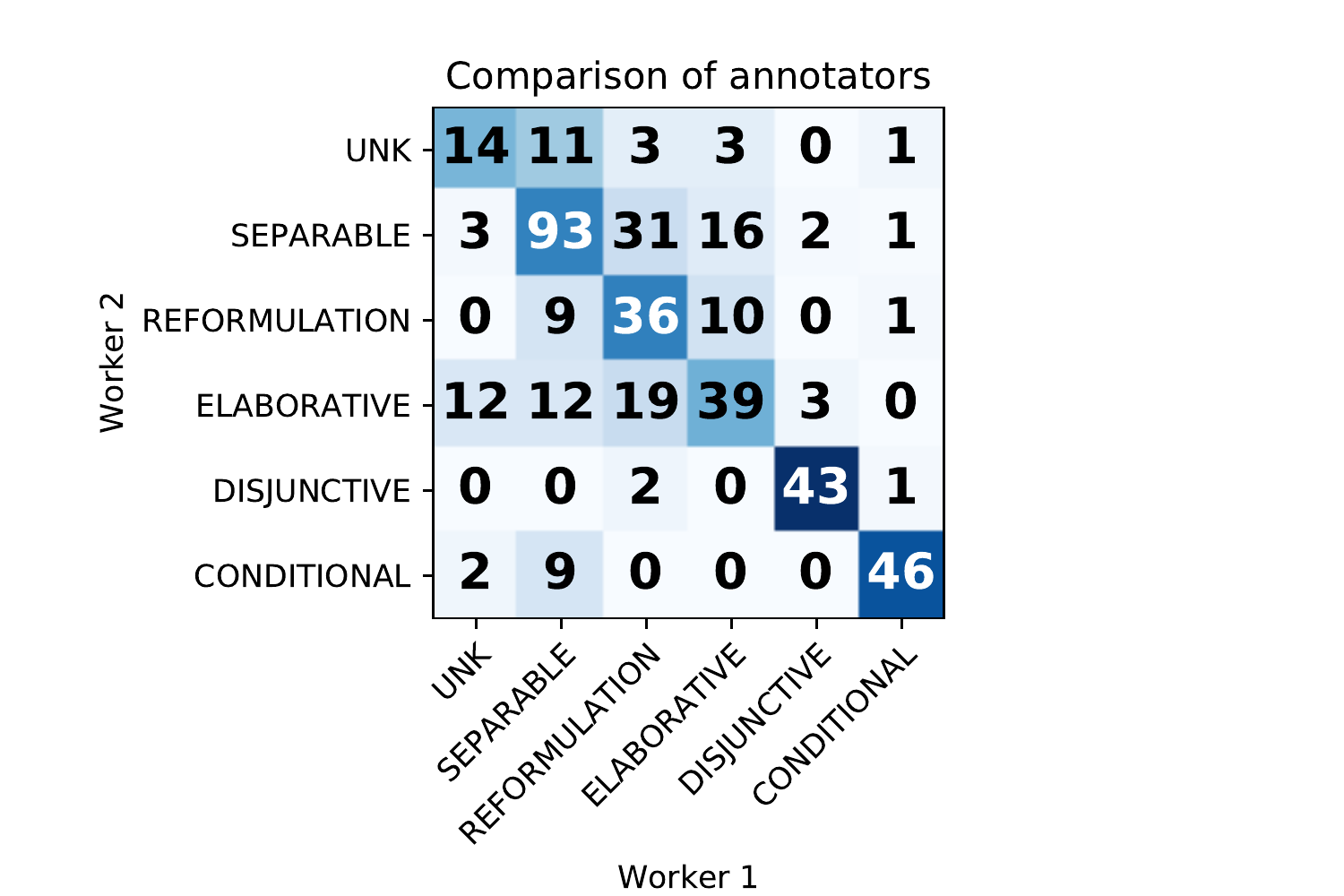}%
    \caption{Confusion matrix between the two annotators who labelled our test set. While there is good agreement on the MSQ types that are often explicitly signalled (\textsc{Disjunctive} and \textsc{Conditional}), the other types are often more subtle, and examples may involve multiple intents.}
        \label{fig:cm_annotators}%
\end{figure}

\end{document}